%% file: sample-sigconf.tex
\documentclass[sigconf]{acmart}
\AtBeginDocument{%
  }

\copyrightyear{2025}
\acmYear{2025}
\setcopyright{cc}
\setcctype{by-sa}
\acmConference[CIKM '25]{Proceedings of the 34th ACM International Conference on Information and Knowledge Management}{November 10--14, 2025}{Seoul, Republic of Korea}
\acmBooktitle{Proceedings of the 34th ACM International Conference on Information and Knowledge Management (CIKM '25), November 10--14, 2025, Seoul, Republic of Korea}\acmDOI{10.1145/3746252.3761305}
\acmISBN{979-8-4007-2040-6/2025/11}

\usepackage{algorithm}
\usepackage{algpseudocode}
\usepackage{makecell}
\usepackage{multirow}
\usepackage{subfig}
\usepackage{graphicx}

\begin{document}

\title{Chunked Data Shapley: A Scalable Dataset Quality Assessment for Machine Learning}


\author{Andreas Loizou}
\affiliation{%
  \institution{\textit{Database and Knowledge Systems Lab} \\
\textit{School of ECE, National Technical University of Athens}\\}
  \city{Athens}
  \country{Greece}}
\email{antreasloizou@mail.ntua.gr}

\author{Dimitrios Tsoumakos}
\affiliation{%
  \institution{\textit{Database and Knowledge Systems Lab} \\
\textit{School of ECE, National Technical University of Athens}\\}
  \city{Athens}
  \country{Greece}}
\email{dtsouma@mail.ntua.gr}








\begin{abstract}
As the volume and diversity of available datasets continue to increase, assessing data quality has become crucial for reliable and efficient Machine Learning analytics. A modern, game-theoretic approach for evaluating data quality is the notion of \textit{Data Shapley} which quantifies the value of individual data points within a dataset. State-of-the-art methods to scale the NP-hard Shapley computation also face severe challenges when applied to large-scale datasets, limiting their practical use. In this work, we present a Data Shapley approach to identify a dataset's high-quality data tuples, \emph{Chunked Data Shapley (C-DaSh)}. \emph{C-DaSh} scalably divides the dataset into manageable chunks and estimates the contribution of each chunk using optimized subset selection and single-iteration stochastic gradient descent. This approach drastically reduces computation time while preserving high quality results. We empirically benchmark our method on diverse real-world classification and regression tasks, demonstrating that \emph{C-DaSh} outperforms existing Shapley approximations in both computational efficiency (achieving speedups between 80$\times$ -- 2300$\times$) and accuracy in detecting low-quality data regions. Our method enables practical measurement of dataset quality on large tabular datasets, supporting both classification and regression pipelines.
\end{abstract}




\received{22 May 2025}
\received[revised]{04 August 2025}
\received[accepted]{27 August 2025}

\maketitle

\section{Introduction}
\input{Sections/Introduction}

\section{Problem Formulation} \label{section2}

\input{Sections/problem_motivation}

\section{Methodology}\label{section3}
\input{Sections/Methods}

\section{Experimental Evaluation}\label{section4}
\input{Sections/experimental_Evaluation}

\section{Related Work}\label{section5}
\input{Sections/related_work}

\section{Conclusions and Future Work}\label{section6}

\input{Sections/conclusion_and_fw}

\section{GenAI Usage Disclosure}
Generative AI tools such as Grammarly and ChatGPT were used in some sections simply to improve the text's flow and address grammar and syntax mistakes.  

\begin{acks}
This work is partially supported by the project RELAX-DN, funded by the European Union under Horizon Europe 2021-2027 Framework Programme Grant Agreement number 101072456.
\end{acks}

\bibliographystyle{ACM-Reference-Format}
\balance
\bibliography{sample-base}

\end{document}

%% file: Sections/Introduction.tex
Data grows exponentially, in volume as well as variety and velocity \cite{hazen2014data}. As data increases at this pace in data centres, organisations struggle with the performance of their decision-making algorithms, one promising reason is the data quality. Selecting the ``right'' data for the used of an data-driven algorithm (such as machine learning) can improve the algorithm's performance efficiency \cite{loizou2025VEOM}. 

The plethora of available data within the organisation's data centres can cause the identification of high-quality data to be a real struggle. Using the human factor to identify high-quality data is time-consuming, as well as drop the performance of the data-driven algorithm. Automating data selection using AI techniques has been gamy tracing the data context is named as Data-Centric Artificial Intelligence (AI) \cite{zha2025data, loizou2025VEOM, loizou2025venom}. Data-Centric AI techniques besides automate the data selection, improve the selection of the high quality data and the performance of a data-driven ML algorithm. Also, because data quality is described on different dimensions using Data-Centric AI can focus on the specific dimension's that matters \cite{jakubik2024data}.

What constitutes \textbf{Data Quality} can vary depending on different contexts and applications. Previous studies have focused on traditional data quality issues, such as low accuracy due to poor data quality, dataset completeness with respect to real-world scenarios, and consistency among the data tuples within a dataset \cite{wang1996beyond, batini2009methodologies}. More recent studies define low data quality with respect to the size, order, and distribution of the data \cite{giannakopoulos2018content}. Others emphasize on more nuanced dimensions, such as bias and fairness, which can result to discriminatory results in machine learning models trained on suboptimal datasets \cite{pitoura2020social, barry2023impact}. A recent study \cite{mohammed2025effects} categorises data quality in six different dimensions: representation, completeness, feature accuracy, target accuracy, uniqueness and target class balance. Data Quality can be depended on the data tuple features, such as corruption or missing, which is a problem that Data Shapley works try to resolve \cite{shapley1953value, roth1988shapley}. As a data-centric AI approach, \textbf{Data Shapley} quantifies the importance of individual data tuples in a data-driven learning algorithm, identifying its quality and value by their contribution to overall model performance \cite{wang2024rethinkingdatashapleydata, zha2025data}. However, computing Data Shapley is an NP-hard problem, and various approximations have been designed to improve its computational efficiency \cite{ghorbani2019data, wang2024data}. Data Shapley approximations remain computational expensive, often require hours to find each data tuple quality value for a whole dataset with thousands of lines working in parallel mode \cite{ghorbani2019data}. In a Data Shapley approximation study \cite{mitchell2022sampling}, they mentioned that a critical component in making these approximations practical is the selection of the most informative subset combinations, which can significantly influence both accuracy and efficiency.

For instance having a repository like MIMIC-III, which contains daily patient reports from Beth Israel Deaconess Medical Center \cite{johnson2018mimic}, it is crucial to identify the missing or corrupted data tuples, as they can negatively impact ML algorithms. Previous studies \cite{ghorbani2019data, mitchell2022sampling}, showed that approximate methods can be used to detect low-quality data tuples. However, these methods still suffer from high computational complexity, often requiring hours to days. This makes their application to large-scale, real-world scenarios impractical.

In order to enable practical assessment of dataset quality, we propose a Data Shapley approximation method called \textit{Chunked Data Shapley (C-DaSh)}. This method aims to close the gap between the computational efficiency and identify the quality of each data tuple accurately. \textit{C-DaSh}, instead of evaluating the contribution of each individual data tuple, groups data tuples into chunks and computes the contribution of each chunk to the performance of a data-driven ML algorithm. By incorporating a subset selection prior to the Data Shapley computation, we identify the most optimal subsets which enhances the ability of our method to identify low-quality data chunks. Unlike most existing Data Shapley approximation methods that are limited to either classification or regression/time-series tasks, our approximation method is designed to work effectively for both tasks. In comparison to prior baseline work by \cite{ghorbani2019data}, which introduces two Data Shapley approximations methods, the Gradient Shapley(G-Shapley) and Truncated Monte Carlo Shapley (TMC-Shapley), our approach demonstrates improved accuracy in detecting low-quality data and achieves considerably improved performance.

The main contributions of our work can be summarised as follows:
\begin{itemize}
    \item We introduce a novel Data Shapley approximation algorithm, \textit{C-DaSh}, that partitions datasets into chunks and computes quality scores per chunk instead of individual data tuples. Using our subset selection process to identify the most informative data chunks, our method effectively reduces computational complexity from exponential to manageable levels with no degradation in accuracy. 
    \item We extend the applicability of our method to support both classification and regression/time-series tasks, demonstrating versatility across different domains.
    \item We provide an experimental evaluation across multiple real-world datasets with varied data quality issues (noise, missing values, label corruption), comparing our method against two established Data Shapley approximation techniques. Results show that \textit{C-DaSh} can identify data quality more accurately, achieving impressive speedups ranging from 80$\times$ to 2300$\times$. We also provides insights into optimal chunk sizes and subset configurations, balancing accuracy and efficiency for real-world deployments.
\end{itemize}

Our work is structured as follows: Section \ref{section2} presents the Data Shapley problem formulation, along with two baseline approximations. Section \ref{section3} outlines the motivation behind our proposed approach, based on the limitations of existing methods and emerging requirements. It also introduces our proposed method, \textit{C-DaSh}, and the adopted subset selection strategy. In section \ref{section4}, we provide a comprehensive experimental evaluation, comparing \textit{C-DaSh} against two baseline methods, and exploring trade-offs between different configuration settings of \textit{C-DaSh}. Related work in the area of data quality, both with and without the Data Shapley approach, is described in Section \ref{section5}. Section \ref{section6} concludes the paper with key findings and directions for future research.

%% file: Sections/problem_motivation.tex
Consider a large collection of tabular datasets without any information related to it. Each dataset $D$ comprises data tuples $x_j$ represented as $D = \left( x_1,\ x_2,\ x_3,\ \ldots,\ x_n\right)$. Each data tuple $x_j$ has the same number of features with the other data tuples, as well as be on the same domain. For a given Machine Learning (ML) algorithm $\mathcal{A}$ (such as Multilayer Perception, SVM, etc.) we aim to assess the ``quality'' of the data tuples and distinguish between the low data quality with the high ones.

Data Shapley \cite{shapley1953value, roth1988shapley}, rooted in game theory, offers a principled way to evaluate the contribution or ``quality'' of each data tuple as a player. Having a performance metric $M$ such as Mean Square Error, Accuracy, F-1 score, etc., and a learning algorithm $\mathcal{A}$, finds for each data tuple its quality value, find its contribution among all the available subset combinations using all the rest data tuples from the dataset $D$. Data Shapley value $ds_j$ for a data tuple $x_j$ and a performance metric $M$ can be computed as:
\begin{equation}
    ds_{x_j}^M = \sum_{Z \subseteq D \setminus \{j\}} \frac{|Z|! \cdot (|D| - |Z| - 1)!}{|D|!} \left(M(Z \cup \{j\}) - M(Z)\right)
\end{equation}
where $Z$ is a subset with data tuples (set of players) from dataset $D$. When a subset $Z$ consists all the rest data tuples except $x_j$ is named as the ``grand coalition''. After the Data Shapley algorithm, is applied on $D$, we have a $DS = \left( ds_1,\ ds_2,\ ds_3,\ \ldots,\ ds_n\right)$, where $n$ is the total number of tuples in dataset $D$. Low $ds_j$ values denote low-quality tuples and vice-versa. However, Data Shapley is an NP-hard problem \cite{deng1994complexity}, with a complexity of $O(2^n)$, making it impossible to identify the quality of a large-scale dataset. 

Truncated Monte Carlo Shapley \cite{ghorbani2019data} is a Data Shapley approximation that tries to speedup the calculation of Data Shapley values. Firstly, this method applies a random permutation on the tuples of dataset $D$. For each Monte Carlo iteration, is applied for each data tuple, the Data Shapley equation to find its marginal contribution on the ML algorithm $\mathcal{A}$. For each data tuple $x_j$ the subset $Z$ consists all the permuted data tuples before the $x_j$, such as $Z =  \left( x_1,\ \ldots,\ x_{j-1}\right)$. Multiple Monte-Carlo permutations with the same procedure are repeated until the \textit{truncation} is met. Truncation is a strategy used to assess whether the performance of a learning algorithm $\mathcal{A}$ on a subset $Z$, as measured by the performance metric $M$, meets a predefined tolerance threshold. It helps determine when the marginal contributions of all data tuples are sufficiently accounted for. The final Data Shapley value is the average from all Monte-Carlo permutations marginal contribution of the specific tuple. However, this approximation may speedup the runtime efficiency, with complexity $O(m\cdot n)$ (in practice $m\approx3n$, so $O(n^2)$) where $m$ is the total number of Monte-Carlo permutations until the truncation is met. However, for a large-scale dataset with thousands of tuples is still not efficient.

Gradient Shapley \cite{ghorbani2019data} is another Data Shapley approximation. Using the Stochastic Gradient Descent (SGD) \cite{amari1993backpropagation} on algorithm $\mathcal{A}$ when is applicable can use the knowledge from the subset data tuples and distinguish the high-quality data tuples from the lower one more accurately. This approximation uses randomly selected batches with random permutations, similarly to the Truncated Monte Carlo Shapley. Using the random subsets update the model hyperparameters. For each data tuple applies the Data Shapley equation using the model hyperparameters information from the SGD which gained it from the previous data tuples. Using that information find for $x_j$, its marginal contribution on the algorithm $\mathcal{A}$ and the performance metric $M$. The subset $Z$ for each data tuple $x_j$ is all the previous data tuples where update the model hyperparameters using SGD, such as $Z =  \left( x_1,\ \ldots,\ x_{j-1}\right)$. The complexity time in this approximation is $O(n\cdot T \cdot d)$, where $d$ is the model hyperparameters dimensionality and $T$ is the number of gradient descent steps (how many times random permutations are performed). This approximation may improve the runtime efficiency compared to the exact Data Shapley computation, but for large-scale datasets will not be efficient.

%% file: Sections/Methods.tex


\subsection{Motivation}

Determining a multi-dimensional ``quality'' score for each data tuple in a large-scale dataset capturing aspects such as noise, completeness, and more is a highly challenging task. It involves addressing various additional complexities and circumstances, including what performance metric $M$ you must couple with learning algorithm $\mathcal{A}$ and how you efficient is the quality value of the identification on the low-quality data tuples \cite{ghorbani2019data}. Data Shapley has the ability to integrate a lot of data quality dimensions, but its high complexity makes it difficult for large-scale datasets. Data Shapley approximation methods have been introduced that try to improve the runtime efficiency, without affecting the identification of the low data quality. However, their practicality is limited to datasets with a ten of thousands of tuples \cite{ghorbani2019data}. Nowadays, datasets comprise hundreds to millions data tuples, making it difficult to apply these approximations as they would require months to execute to find the ``right'' high-quality data tuples for an ML algorithm. Our motivation stems from these limitations. We aim to design and implement a novel approximation of the Data Shapley that not only scales to large datasets but also provides an interpretable, high-precision distinction between the low-quality data tuples and the high-quality among a large dataset. Computing per tuple Data Shapley values is computationally expensive, our approach leverages chunk-based strategy assigning a quality value to each chunk. Our approximation method incorporates SGD \cite{amari1993backpropagation}, to more accurately estimate the contribution of each chunk. Additionally, we employ an smart subset selection strategy to enhance the identification of low quality chunks. Together, these techniques significantly reduce computational overhead while preserving the integrity and precision of the data quality assessment.

\subsection{Chunked Data Shapley}

\subsubsection{Data Quality Measurement}
To reduce the computational complexity of calculating the Data Shapley value for an entire dataset $D$, we partition the dataset into $c$ chunks of the same size $l$, where the total amount of chunks $c \ll n$, as well as the chunk size $l \ll n$. For each chunk $ch_j$ from the set $\left( ch_1, ch_2, ch_3, \ldots, ch_c \right)$ we assign a corresponding data quality score value $ds_j$. Upon completing the approximation process for all chunks, we obtain a Data Shapley quality score vector $DS = \left( ds_1,\ ds_2,\ ds_3,\ \ldots,\ ds_c\right)$. Organising the dataset into distinct chunks, non-overlapping multiple data tuples, we significantly speed up computation, as the number of required subset evaluations is greatly reduced. Before computing the Data Shapley values, we perform a subset selection step to identify the most informative and representative subsets $\mathcal{S}$ for evaluation. As highlighted in \cite{mitchell2022sampling}, careful subset selection leads to more accurate quality estimates for individual data tuples. A detailed explanation of our subset selection strategy is provided in Subsection~\ref{subset-selection-subsection}.

\input{algorithms/algorithm1}

First we partition the dataset $D$ into chunks with equal size of $l$. Once the subset selection step (described at subsection \ref{subset-selection-subsection}) is performed, we apply the Data Shapley computation over the selected subsets of chunks by using the knowledge information from the SGD \cite{amari1993backpropagation} optimisation function. We use the SGD optimisation function to retrieve from the model hyperparameters the chunks data tuples knowledge information and identify their true quality value score more accurately. Previous Data Shapley studies \cite{ghorbani2019data, wang2024data} using SGD achieved that and identify the low-data quality and the high one more accurately. For each chunk $ch_j$, we compute its data quality value by considering all subsets $Z \in \mathcal{S}$ that do not contain chunk $ch_j$. Using the subsets $Z$, we create a model checkpoint $w_j$ for chunk $ch_j$, which includes all the data tuples informations from $Z$. One complete single run of our method results in a set of model checkpoints $W = (w_1, w_2,\cdots,w_c)$, where $c$ is the total number chunks. The Shapley value calculation for chunk $ch_j$, over a performance metric function $M$ is calculated as follows:
\begin{equation}   
    ds_{ch_j}^M = \sum_{Z \subseteq \mathcal{S} \setminus \{j\}} \frac{C}{|D| \cdot |Z|} \left(M(w_j(Z \cup \{j\})) - M(w_j(Z))\right) \label{eq:c-dash-eq}
\end{equation}
where for the selected subsets $Z \subseteq \mathcal{S}$ do not contain data tuples from chunk $ch_j$ are used to calculate its quality value. When chunk $ch_j$ is not included in any subset, we use all the subsets from $\mathcal{S}$. First we obtain the data information knowledge from $Z$ from the performance metric $M$, and after that we update the model hyperparameters and create the new model checkpoint $w_j$ for chunk $ch_j$ and get its performance for metric $M$. The model checkpoint is calculated using the previous checkpoint $w_{j-1}$ and the optimised loss function using SGD $\mathcal{L}$, for all the chunks in subset $Z$ as:
\begin{equation}
    w_j(Z) = w_{j-1} - \eta_j \sum_{x \in Z} \nabla \mathcal{L}(w_j, x)
\end{equation}
where $\eta_j$ is the learning rate at chunk $ch_j$.
This approach, using the model checkpoint information from all the previous data tuples, optimises the knowledge of the current chunk of data tuples to identify if it contains low-quality data tuples. This approximation with the combination of the optimal subset selection leads to an accurate Shapley value and identifies more easily the data chunks that contain low-quality data tuples, as well as require fewer iterations for computation the Shapley value. Similarly to the existing approaches a lower value for $ds_j$ means less valuable is for the ML algorithm $\mathcal{A}$. Algorithm \ref{algo_1_Less_ICDS} outlines the complete procedure, taking as input a collection a dataset with $n$ tuples, a machine learning algorithm $\mathcal{A}$, a performance metric $M$, and the chunk size $l$. The procedure is repeated until the truncation criterion is satisfied, following a similar approach to the work of \cite{ghorbani2019data}. 

\subsubsection{Subset Selection} \label{subset-selection-subsection}

\input{algorithms/algorithm-subset-select}
For Data Shapley approximations to come closer to the actual value of the original algorithm, it is essential to select an ``optimal'' amount of subsets. Initially, $k$ subsets are randomly chosen, ensuring that each new subset contains data points not present in more than twenty-five percent of the previously selected subsets \cite{mitchell2022sampling}. The number of subsets $k$ is much smaller than the number of chunks $k<<c$, as well as the length of the dataset ($k<<n$). This reduces redundancy and helps the algorithm better capture diverse data contributions. To ensure that we select the subset with the highest quality of data tuples, using the algorithm $\mathcal{A}$ over all the subsets $\mathcal{S}$, we capture for each subset how well it performs over the metric $M$. Then, for each subset $j$ where his metric from the function $M$ is lower than the threshold $\textit{th}$ for classification task and higher for regression, we make the creation of the subset until achieved better performance than the threshold $\textit{th}$. Algorithm \ref{algo_2_Subset_Selection}, depict the subset selection method taking as input all the chunks $ch$, a machine learning algorithm $\mathcal{A}$, a performance metric $M$, and a threshold $\mathit{th}$.

%% file: algorithms/algorithm1.tex
\begin{algorithm}[t!]
\caption{Chunked Data Shapley}
\label{algo_1_Less_ICDS}
\begin{algorithmic}[1]
\Require Dataset $D$, Machine Learning algorithm $\mathcal{A}$, performance metric $M$, chunk size $l$, total number of subsets $k$, threshold $th$, Constant variable $C$
\State \textbf{Initialize} Split dataset $\mathbf{D}$ into $\mathbf{c}$ chunks of size $\mathbf{l}$
\State \textbf{Initialize} Chunks Data Shapley quality values $\mathbf{DS}$ as $ds_j = 0$ for $j = 1, .... , c $
\While{Truncation Criterion is not Met}
\State \textbf{Initialize} $\mathbf{k}$ Subsets from dataset $\mathbf{D}$
\State \textbf{Execute} subset selection on $\mathcal{S}$ using threshold \textit{$\mathbf{th}$}
\State \textbf{Initialize} $\mathbf{w_0} \leftarrow $ Random Parameters

\For{each chunk $ch_j \in D$ for $j = 1, ..., c$}
    \State \textbf{Include} Subsets $Z \subseteq \mathcal{S}\setminus\{j\}$
     \State $w_j(Z) = w_{j-1} - \eta_j \sum_{x \in Z} \nabla \mathcal{L}(w_j, x)$
      \State $\scriptstyle{ds_{ch_j}^M = \sum_{Z \subseteq \mathcal{S} \setminus \{j\}} \frac{C}{|D| \cdot |Z|} \left(M(w_j(Z \cup \{j\})) - M(w_j(Z))\right) \label{eq:ds-eq}}$      
\EndFor
\EndWhile

\State \textbf{Having} Chunk Data Shapley Quality values $DS = \left( ds_1,\ ds_2,\ ds_3,\ \ldots,\ ds_l\right)$
\State \textbf{Return} Dataset quality values $DS$
\end{algorithmic}
\end{algorithm}

%% file: algorithms/algorithm-subset-select.tex
\begin{algorithm}[t!]
\caption{Subset Selection}
\label{algo_2_Subset_Selection}
\begin{algorithmic}[1]
\Require Chunks $ch$, total number of subsets $k$, Machine Learning algorithm $\mathcal{A}$, performance metric $M$, threshold $\mathit{th}$
\State \textbf{Initialize} randomly $k$ subsets $\mathcal{S}$ 

\For{each chunk $ch_j \in D$ for $j = 1, ..., c$}
    \While{$ch_j$ is more than $25\%$ in $Z\in\mathcal{S}$}
        \State \textbf{Initialize} new subsets $Z$ for the $\mathcal{S}$
    \EndWhile
\EndFor

\For{each Subset $Z_j \in \mathcal{S}$ for $j = 1, ..., k$}
    \State \textbf{Get} $M_j$ from $\mathcal{A}$ using Subset $Z_j$
    \While{$M_j$ perform worst than the $\mathit{th}$}
        \State \textbf{Initialize} new Subset $Z_j$ for the $\mathcal{S}$
        \State \textbf{Check} that none $ch$ is appear more than $25\%$ in $Z\in\mathcal{S}$
        \State \textbf{Get} $M_j$ from $\mathcal{A}$ using new subset $Z_j$
    \EndWhile
\EndFor

\State \textbf{Return} Subsets $\mathcal{S}$

\end{algorithmic}
\end{algorithm}

%% file: Sections/experimental_Evaluation.tex
\begin{figure*}[!t]
        \subfloat[Adult Dataset]{
        \includegraphics[width=0.3\textwidth]{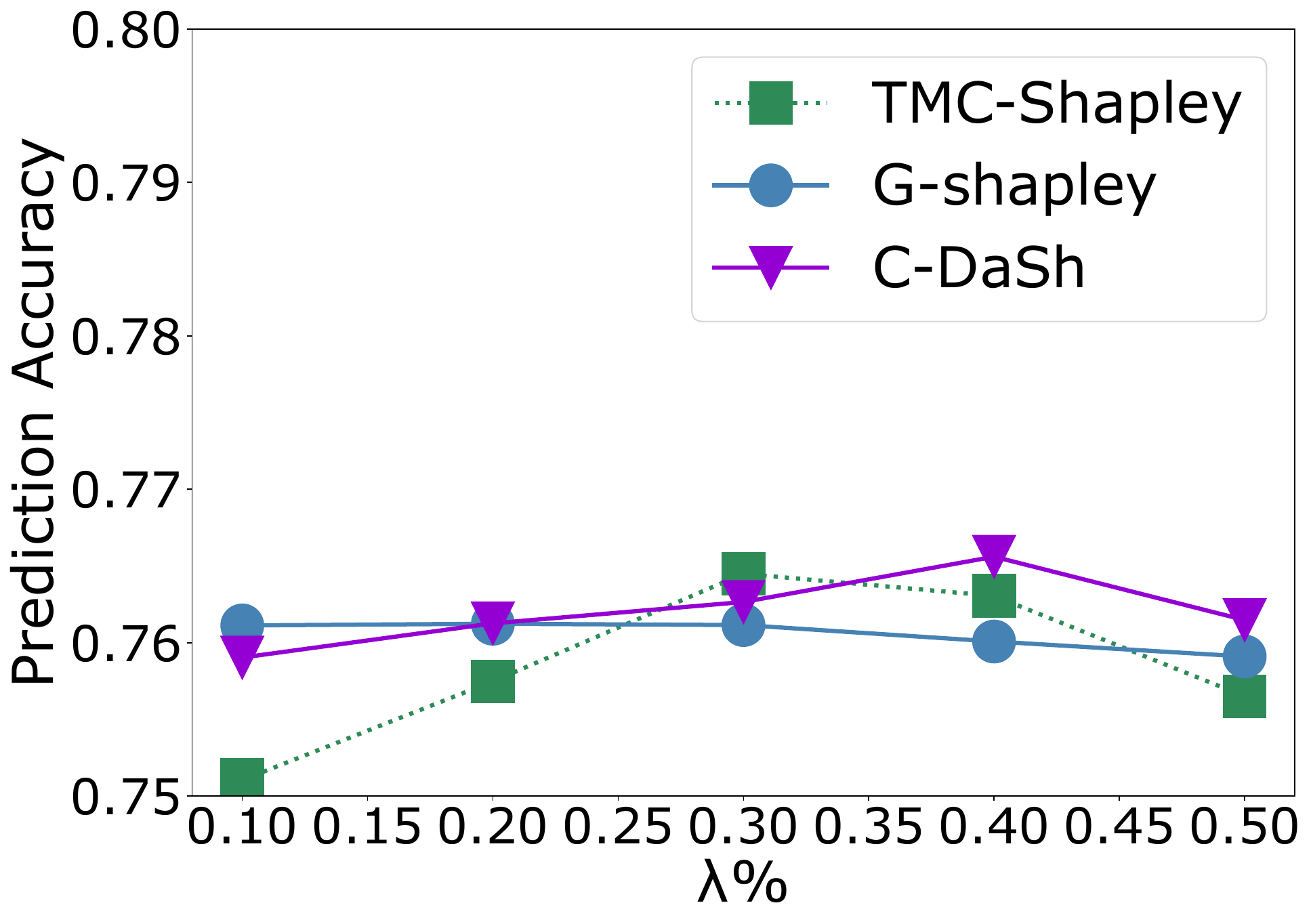}
        \label{fig:eval1-Adult}
    }
    \subfloat[Bank Marketing Dataset]{
        \includegraphics[width=0.3\textwidth]{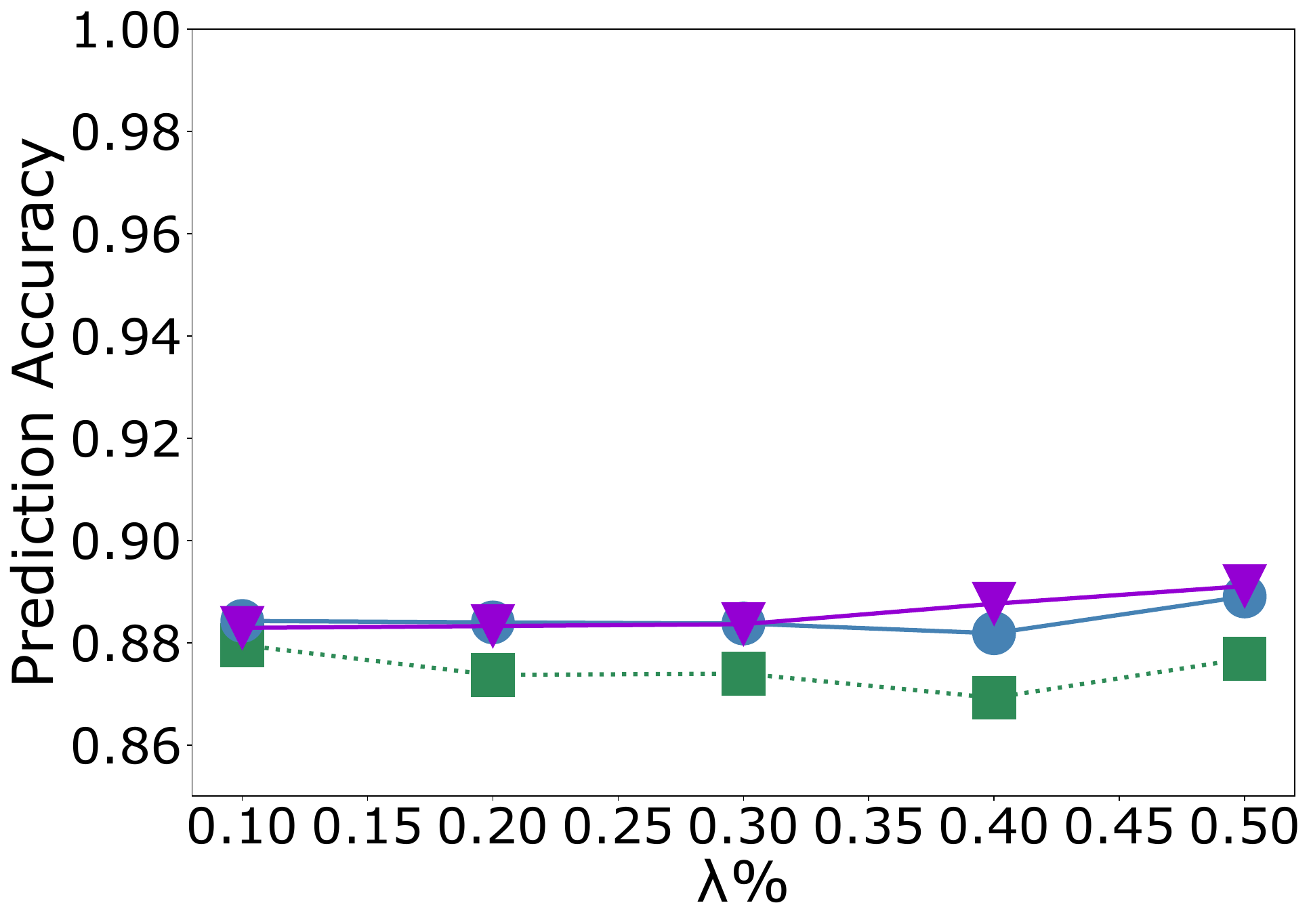}
        \label{fig:eval1-Bank}
    }
    \subfloat[MIMIC-III Dataset]{
        \includegraphics[width=0.3\textwidth]{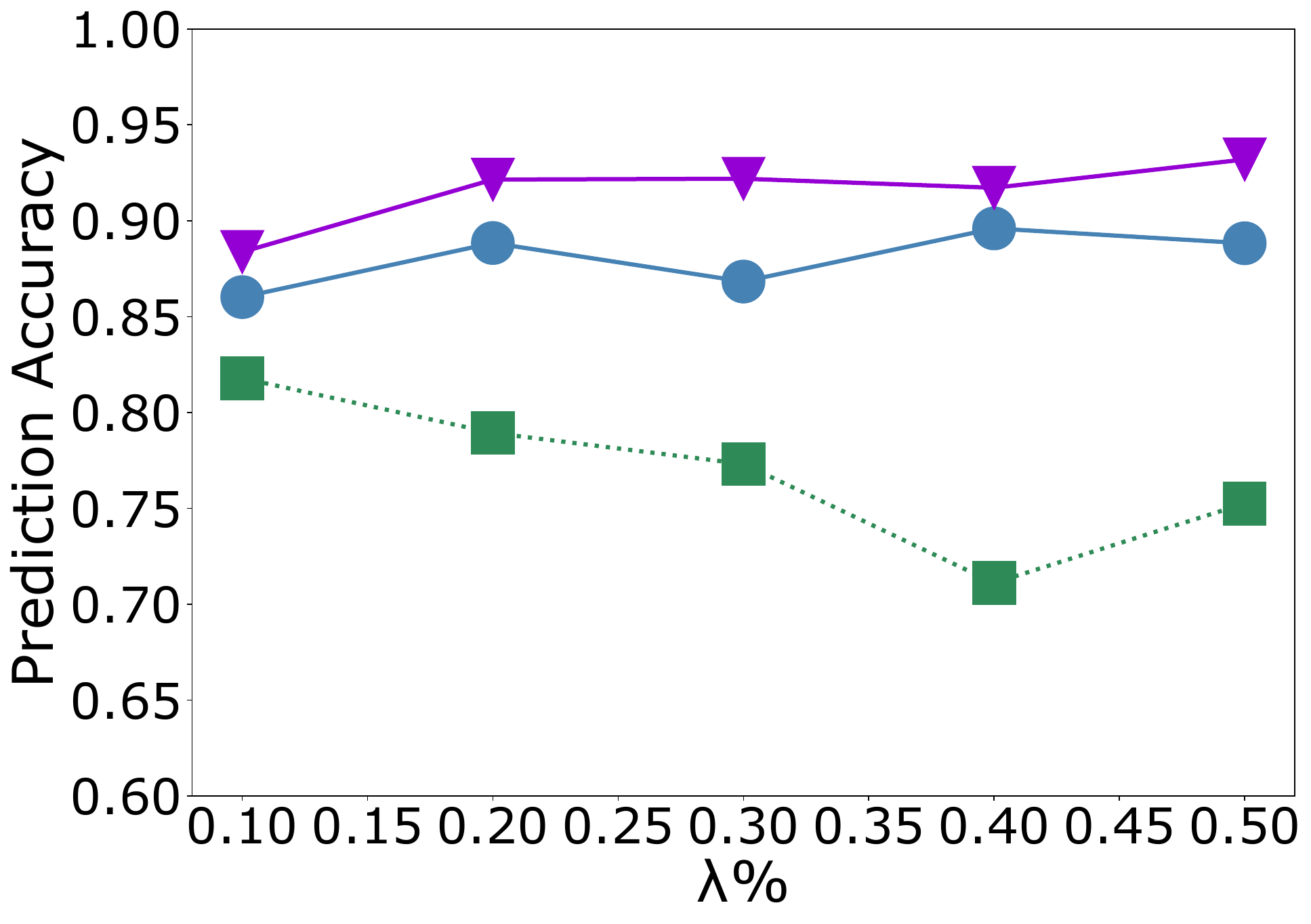}
         \label{fig:eval1-Mimic}
    }
    \caption{Compare C-DaSh with G-Shapley and TMC-Shapley prediction accuracy by removing low data quality from datasets for classification task.}
    \label{fig:eval1}
\end{figure*}


Our proposed Data Shapley approximation method, \textit{C-DaSh}, is evaluated against two baseline approximation algorithms from \cite{ghorbani2019data}, available in an open-source repository. For all experiments involving a classification task, we employ a Multi-Layer Perceptron (MLP) \cite{almeida2020multilayer, popescu2009backpropagation, rosenblatt1958perceptron} with two hidden layers, and for the regression task experiments, a similarly structured MLP tailored accordingly. The learning rate $\eta$ for all methods using SGD was set to $0.001$. The parameters settings for the approximation algorithms are the same as those used in \cite{ghorbani2019data}. In our work, the setting for the threshold \textbf{th} for the subset selection has been set at $0.5$ for the classification experiments and $25$ for the regression experiments. The threshold for classification was selected because it corresponds to the mean of the accuracy range ($0$ to $1$). For regression, the threshold was chosen the average RMSE after computed the algorithm $\mathcal{A}$ using the entire training dataset. First, we discuss the efficiency of removing low-quality data tuples from the datasets and compare MLP’s prediction accuracy. These low-quality data tuples exhibit different such us noise, label corruption, missing data, etc. To evaluate prediction performance, we train the MLP model five times for each experiment, recording the prediction accuracy and reporting the average result. Using the Local Outlier Factor (LOF), we demonstrate the efficiency of our method in identifying outliers in large scale datasets. Therefore, we compare the computational efficiency of our approximation algorithm with the baseline methods from \cite{ghorbani2019data}. Relative speedup was calculated using the formula $S = \frac{T_1}{T_2}$, where $T_1$ is the execution time of the baseline method and $T_2$ is the execution time of \textit{C-DaSh} with the optimal subset selection. We demonstrate the effectiveness of \textit{C-DaSh} through its ability to identify low-quality chunks in regression datasets where the temporal order plays a critical role.

All experiments were executed in a AWS EC2 server with 48 vCPUs of AMD EPYC 7R32 processors at 2.40GHz, and four A10s GPUs with 24GB of memory each, $192GB$ of RAM memory, and $1.5TB$ of storage, running Ubuntu 24.4 LTS. 

\subsection{Datasets}

\begin{table}[!ht]
    \centering
    \setlength\doublerulesep{0.5pt}
    \caption{Dataset properties for experimental evaluation}
    \label{tab:table-evaluation-datasets}
    \begin{tabular}{||c|c|c|c||}
        \hline
         \makecell{Algorithm}& \makecell{Dataset Name} & \makecell{\# Files} & \makecell{\# Tuples}\\ \hline\hline
          \multirow{3}{*}{\makecell{\\Classification}}& \makecell{Adult \cite{AdultDataset}} & $1$ & $40K$ \\ \cline{2-4}
          & \makecell{Bank \\Marketing \cite{bank_marketing_222}} & 1 & $60k$ \\ \cline{2-4}
          & \makecell{MIMIC-III \cite{johnson2018mimic}} & 1 & $55k$ \\ \cline{1-4} 
          \multirow{2}{*}{\makecell{\\\\Regression}}& \makecell{Air Quality \cite{beijing_multi-site_air_quality_501}} & $12$ & $35K$ \\ \cline{2-4}
          & \makecell{Household \\ Power\\ Consumption \cite{HPCdataset}} & 5 & $22k$ - $52k$ \\ \cline{1-4}
        
    \end{tabular}
\end{table}

We evaluated \textit{C-DaSh} using five real-world datasets: Three for classification tasks and two for regression tasks. The Adult \cite{AdultDataset} dataset contains approximately forty thousand individuals with socio-economic features, and poses a binary classification problem: Predicting whether an individual’s income exceeds a certain amount. It contains a substantial number of low-quality data tuples, often affected by missing or corrupted values. The Bank Marketing \cite{bank_marketing_222} dataset consists of data from a Portuguese banking institution, collected through direct marketing campaigns conducted via phone calls. It includes around sixty thousand individuals in a single file, with the classification goal of predicting whether a client will subscribe to a term deposit. The issue with the Bank dataset is that there are a lot of outliers in the data tuples. The MIMIC-III dataset, from Beth Israel Deaconess Medical Centre, is one of the largest available medical datasets, with over forty thousand patients from $2001$ to $2012$. After preprocessing, we created a single dataset to classify whether a patient has sepsis, following an approach similar to \cite{stylianides2025ai}. As noted in \cite{stylianides2025ai}, the MIMIC-III dataset suffers from missing data tuples. The Air Quality \cite{beijing_multi-site_air_quality_501} dataset comprises twelve separate datasets, each from a different nationally controlled air-quality monitoring site. Each dataset contains hourly air pollutants. The regression task is to predict the air quality. The primary data quality issue of this dataset is the high volume of missing data tuples. The Household Power Consumption (HPC) \cite{HPCdataset} dataset contains electric power usage measurements from a household in Sceaux, France. It includes five yearly datasets, each with between $22k$ and $52k$ data tuples recorded at one-minute intervals. The data quality problem in this dataset is the presence of missing values during certain periods of the day. Categorical columns across all datasets were converted to numerical format using one-hot encoding.

\subsection{Baseline Comparison}
\subsubsection{Prediction accuracy and outlier detection evaluation}

Figure \ref{fig:eval1} visualises the comparison of our proposed approximation to the G-Shapley and TMC-Shapley approximations on three classification datasets, since these two baselines apply only to classification tasks.
In this experiment, we obtained the data Shapley values, and using \textit{C-DaSh} we remove the  bottom $\lambda\%$ (ranging from $0.1$ to $0.5$) of data chunks based on their Data Shapley values,  where each chunk has a size of 256. The number of subsets $k$ used in our approximation method was set to $50$. For the G-Shapley and TMC-Shapley the $\lambda\%$ lowest Shapley values of individual data tuples are removed. Figure \ref{fig:eval1} for the three datasets depicts the accuracy (ranging from $0$ to $1$, where higher is better) as a performance metric $M$ of the ML algorithm $\mathcal{A}$ on the y-axis. On the axis-x, the percentage of removed data based on the lowest Shapley values is depicted.

Figure \ref{fig:eval1-Adult} shows that for the Adult dataset, our proposed method, \textit{C-DaSh} has a low improvement impact in prediction accuracy as some of low-quality data are removed. It achieves slightly higher accuracy compared to both G-Shapley and TMC-Shapley across all levels of data removal. For example, with just $10\%$ of the lowest-quality data tuples removed, our method have $1\%$ higher accuracy than TMC-Shapley and almost the same prediction accuracy with G-Shapley. This small improvement stems from optimal subset selection, which allows our method to effectively target the low-quality chunks while preserving high-quality data. A drop in performance typically indicates that most low-quality data tuples has already been eliminated, and further removal begins to affect chunks which containing high-quality data tuples.

Figure \ref{fig:eval1-Bank} illustrates that, for the Bank marketing dataset TMC-Shapley for $\lambda$ equal to or lower than $40\%$, the accuracy slightly drops, suggesting that these methods mistakenly identify high-quality tuples as low-quality. In contrast, \textit{C-DaSh}, as well as G-Shapley has a slight improvement until $30\%$ has same performance without any up-down trend. That's show that both high-quality and low-quality data tuples are removes. For $\lambda$ value higher or equal with $40\%$, \textit{C-DaSh} has higher prediction accuracy with a small difference. This further demonstrates that optimal subset selection assists our method minimise the removal of valuable data while effectively filtering out low-quality data, thereby preserving the training performance of algorithm $\mathcal{A}$.

Figure \ref{fig:eval1-Mimic} demonstrates that, for the MIMIC-III dataset, \textit{C-DaSh} surpass both G-Shapley and TMC-Shapley in accuracy by approximately $7\%$ and $15\%$, respectively. TMC-Shapley exhibits a significant drop in performance accuracy up to $\lambda$ equal with $40\%$, indicating that it misclassifies high-quality tuples as low-quality. G-Shapley performs better in overall, however for $\lambda$ between $20\%$ and $30\%$ purges high-quality data features, while at higher $\lambda$, it begins to identify low-quality data tuples more effectively. \textit{C-DaSh} consistently achieves higher accuracy than both baselines due to its optimal subset selection, which helps identify low-quality data chunks and retain high-quality ones for training algorithm $\mathcal{A}$ across all $\lambda$ values.

\begin{figure}[!t]
    \subfloat[Gaussian Noise as low data quality]{
        \includegraphics[width=0.3\textwidth]{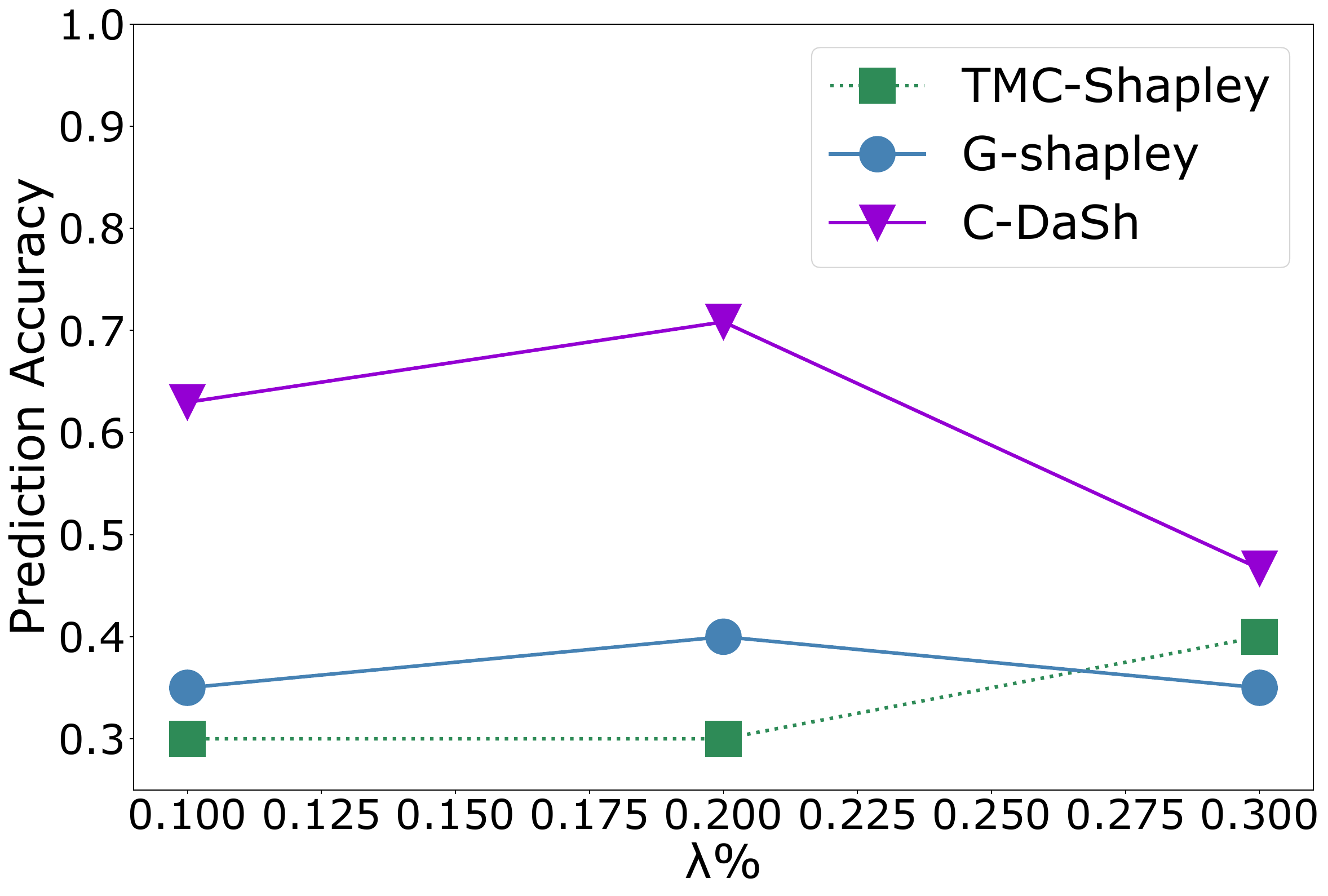}
        \label{fig:eval2-Noise}
    }
    
    \subfloat[Label Corruption as low data quality]{
        \includegraphics[width=0.3\textwidth]{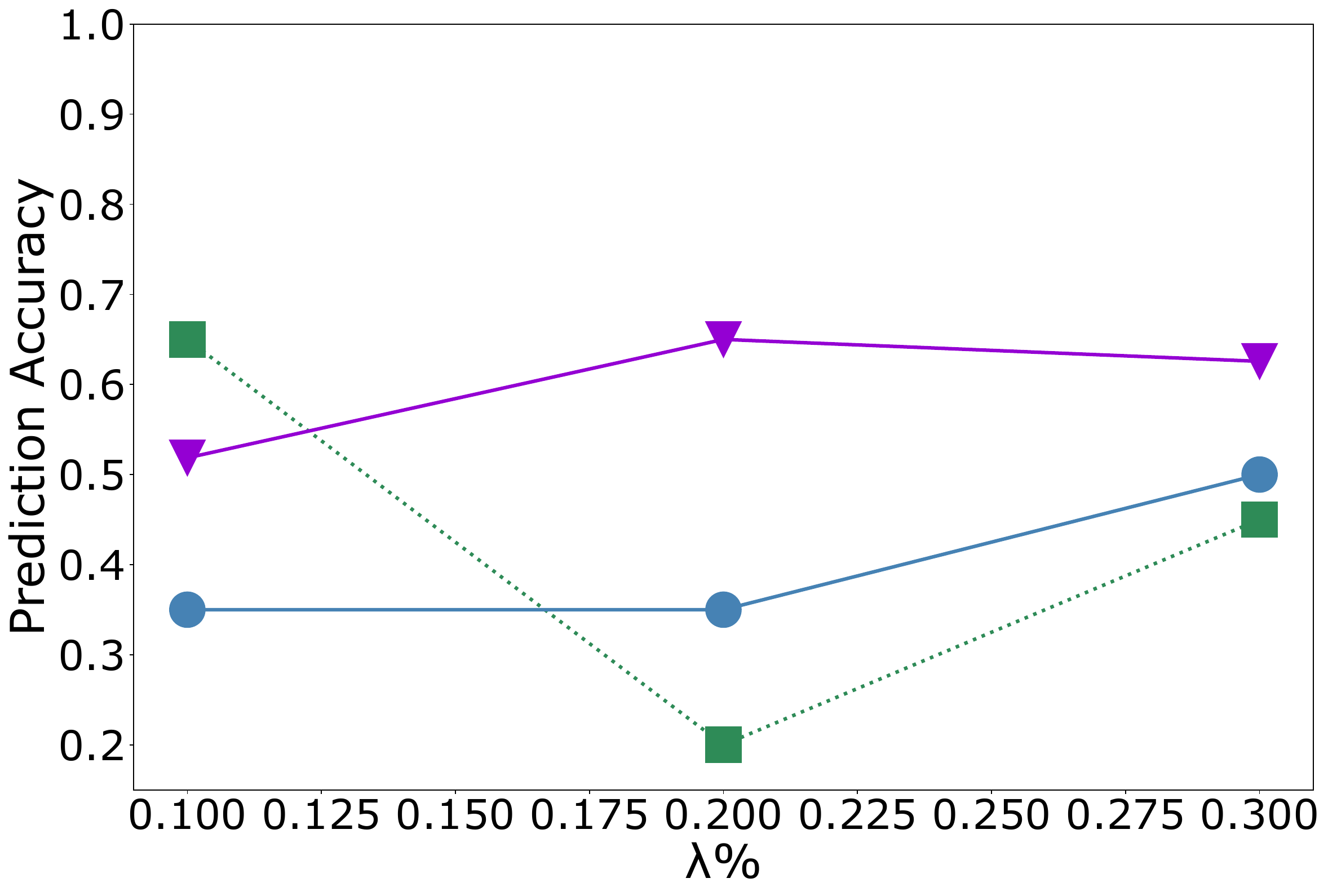}
         \label{fig:eval2-flipped}
    }
    \caption{Compare C-DaSh with G-Shapley and TMC-Shapley prediction accuracy on Adult Dataset with manually annotated different low data quality issues.}
    \label{fig:eval2}
\end{figure}

Figure \ref{fig:eval2} illustrates two different low-quality data scenarios for the Adult dataset, Gaussian noise added to $20\%$ of the data tuples (Figure \ref{fig:eval2-Noise}) and label corruption where $20\%$ of the labels were flipped (Figure \ref{fig:eval2-flipped}). In both experiments, our goal is to exclude data tuples with missing values for better evaluation approximation algorithms' ability to identify and handle specific data quality issues. In both Figures, the x-axis represent the percentage ($\lambda\%$ ) of removed data tuples, while the y-axis reports the prediction accuracy of the learning algorithm $\mathcal{A}$. At $\lambda$ equal to $20\%$, \textit{C-DaSh} successfully removes most of the low-quality data tuples and outperforms both G-Shapley and TMC-Shapley by $30\%$ and $40\%$ respectively. However, G-Shapley identifies some of the high-quality data tuples until  $\lambda$ equals $20\%$, but for higher $\lambda$ values mistakenly classifies some high-quality data tuples as low-quality, and that leads to a significant drop in prediction accuracy. In the label corruption from Figure \ref{fig:eval2-flipped}, TMC-Shapley identifies more low data quality data tuples at $\lambda$ equals with $10\%$ and outperforms \textit{C-DaSh} by almost $12\%$. However, for higher $\lambda$ values, \textit{C-DaSh} consistently outperforms both methods.

\begin{figure}[!t]
    \subfloat[Adult Dataset]{
        \includegraphics[width=0.3\textwidth]{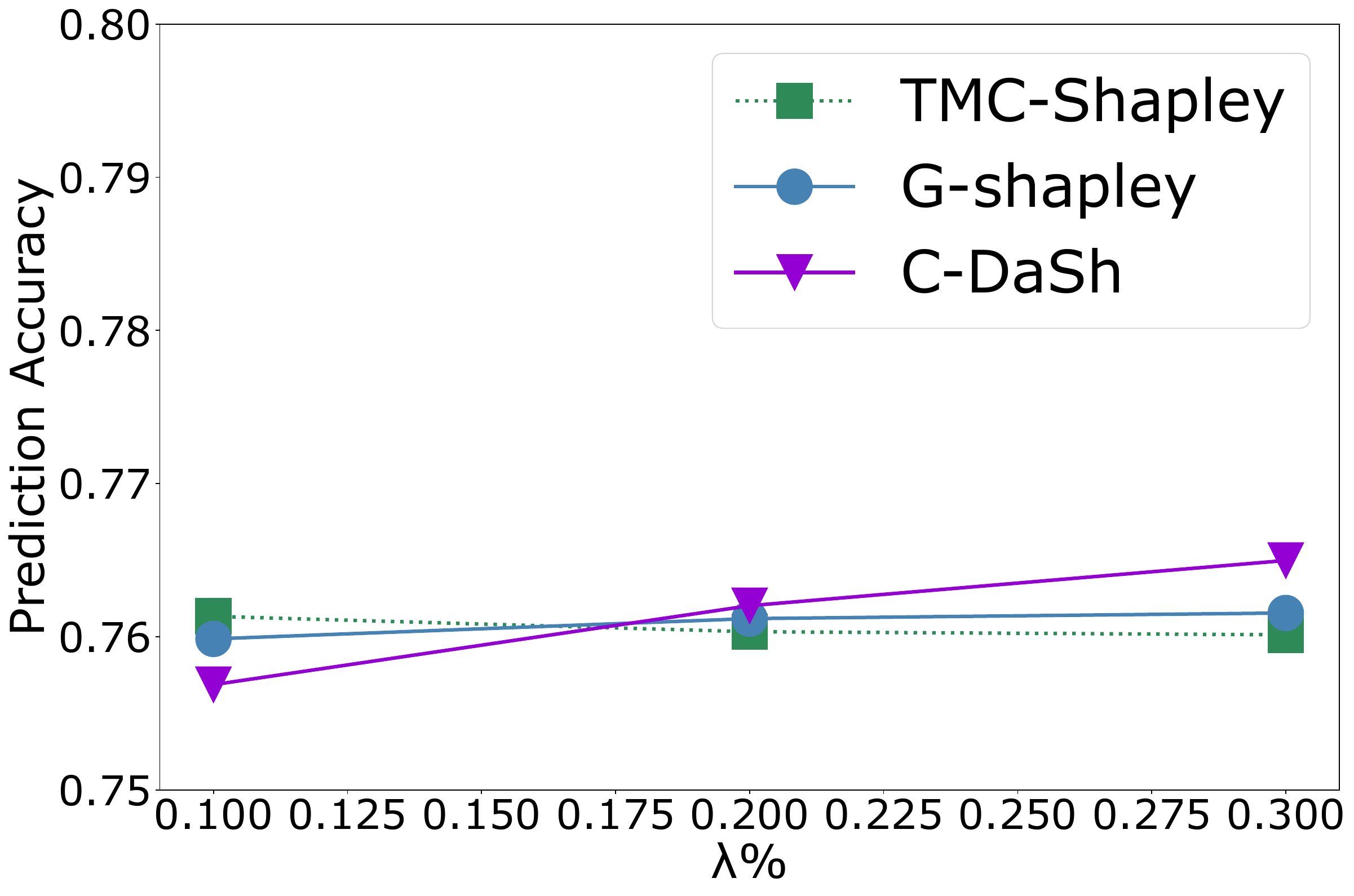}
        \label{fig:eval8-Adult}
    }

    \subfloat[MIMIC-III Dataset]{
        \includegraphics[width=0.3\textwidth]{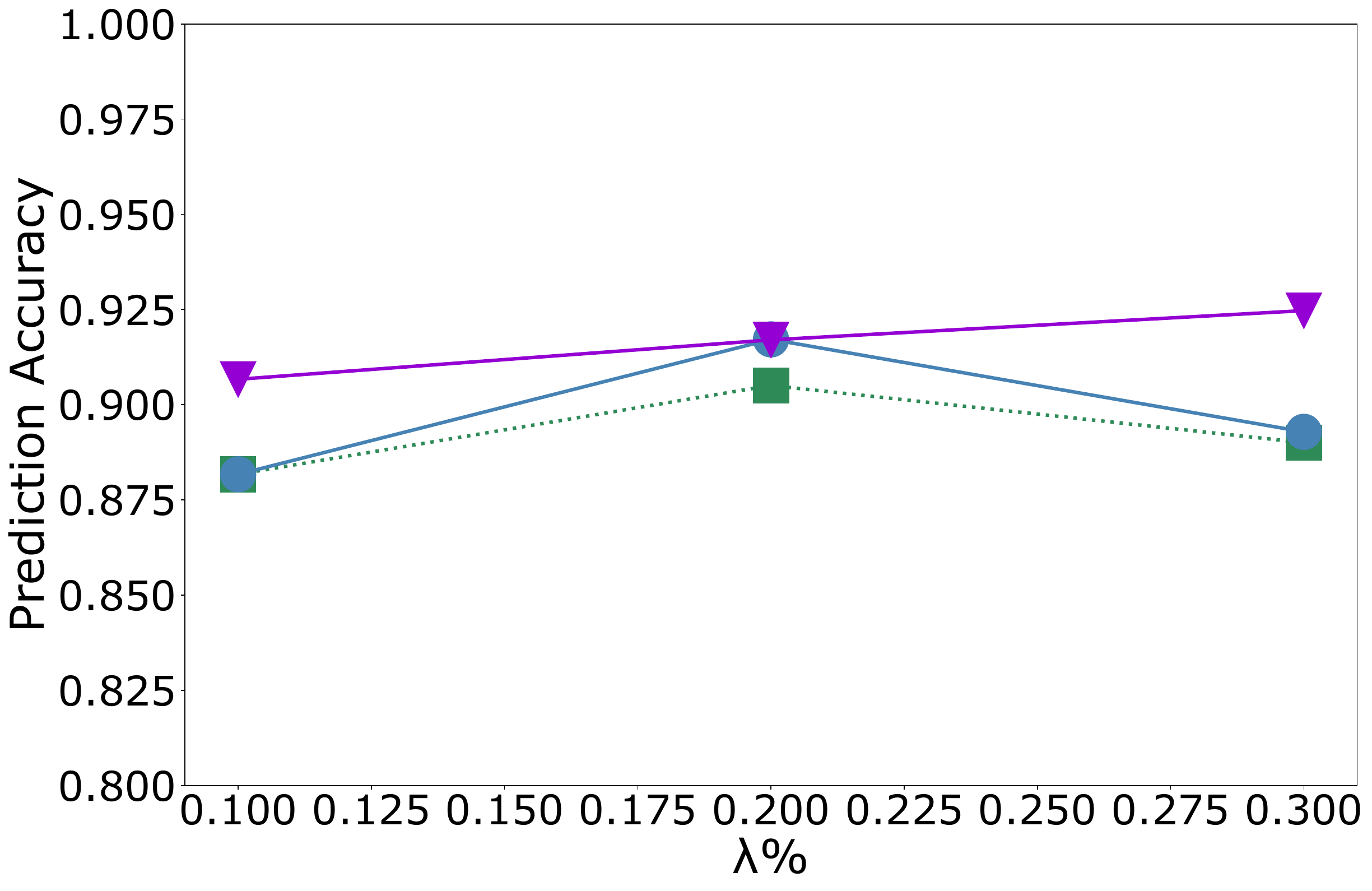}
         \label{fig:eval8-Mimic}
    }
    \caption{Compare prediction accuracy between C-DaSh and a Chunk Average method where uses the quality values from G-Shapley or TMC-Shapley on Adult and MIMIC Datasets by remove the $\lambda\%$.}
    \label{fig:eval8}
\end{figure}

We designed a second method in which we chunk the datasets evaluate the ability of chunking as a method, using either the G-Shapley or TMC-Shapley approximation. The overall quality value of a chunk is computed by averaging for each chunk all of its tuple Shapley scores. As in our proposed \textit{C-DaSh} method, each chunk is assigned a quality value based on one of the two approximation methods. To evaluate how accurately  \textit{C-DaSh} identifies chunk quality score value, we compare it with the chunked version of both approximation methods. All the three methods use the same chunks containing equal number of data tuples (chunk size of $250$ data tuples for the Adult and MIMIC-III datasets). Figure \ref{fig:eval8} presents the prediction accuracy results from these experiments. The y-axis of both sub-figures depicts the prediction accuracy, while the x-axis indicates the percentage of removed low-quality chunks (ranging from $10\%$ to $30\%$). For the Adult dataset (Figure \ref{fig:eval8-Adult}), when $10\%$ of the chunks were removed, both baseline methods slightly outperformed \textit{C-DaSh}, with less than a $0.5\%$ improvement. Furthermore, at higher removal percentages, \textit{C-DaSh} outperformed the chunk-averaging with the data quality values based on the baselines' approximations. This indicates that \textit{C-DaSh} can more precisely identify low-quality chunks with more precision compare to the baseline averaging method. For the MIMIC-III dataset (Figure \ref{fig:eval8-Mimic}), \textit{C-DaSh} consistently achieved higher accuracy than the chunk averaging using the TMC-Shapley across all removal percentages. While G-Shapley at $\lambda$ equals with $20\%$, achieved prediction accuracy comparable to \textit{C-DaSh}. In contrast, \textit{C-DaSh} continued to identify low-quality chunks more effectively. However, both of baseline chunking methods  either performs similarly (G-Shapley), or better (TMC-Shapley) than \textit{C-DaSh} on both datasets. This demonstrates that employing chunking is beneficial for the dataset, as it enables more precise identification of low-quality data chunks.

\begin{table}[!ht]
    \centering
    \setlength\doublerulesep{0.5pt}
    \caption{Data Shapley Approximations Outlier Detection Average Error over $\lambda\%$ percentage of removing low data quality data tuples}
    \label{tab:table-outlier}
    \begin{tabular}{||c|c|c|c|c||}
        \hline
         \makecell{Dataset\\Name}& \makecell{DS\\Approximation\\Method} & \makecell{$\lambda = 0.1$} & \makecell{$\lambda = 0.2$}& \makecell{$\lambda = 0.3$}\\ \hline\hline
          \multirow{3}{*}{Adult}& \makecell{C-DaSh} & $\mathbf{1.260}$ & $1.262$ & $1.266$\\ \cline{2-5}
          & \makecell{TMC-Shapley} & $1.274$ & $1.312$& $1.327$ \\ \cline{2-5}
          & \makecell{G-Shapley} & $1.275$ & $1.312$& $1.327$\\ \cline{1-5} 
          \multirow{3}{*}{\makecell{Bank\\Marketing}}& \makecell{C-DaSh} & $\mathbf{1.046}$ & $\mathbf{1.046}$ & $1.049$\\ \cline{2-5}
          & \makecell{TMC-Shapley} & $1.096$ & $1.076$& $1.068$ \\ \cline{2-5}
          & \makecell{G-Shapley} & $1.096$ & $1.076$ & $1.068$ \\ \cline{1-5} 
          \multirow{3}{*}{MIMIC-III}& \makecell{C-DaSh} & $\mathbf{1.148}$ & $1.150$ & $1.153$\\ \cline{2-5}
          & \makecell{TMC-Shapley} & $1.348$ & $1.340$& $1.303$ \\ \cline{2-5}
          & \makecell{G-Shapley} & $1.348$ & $1.340$& $1.303$ \\ \cline{1-5} 
        
    \end{tabular}
\end{table}

Table \ref{tab:table-outlier} presents the absolute outlier factor scores obtained using the Local Outlier Factor (LOF) \cite{breunig2000lof} method for the three classification datasets of Adult, Bank Marketing, and MIMIC-III. 
LOF detects anomalies by comparing a point’s local density to that of its neighbours, flagging points with significantly lower density as outliers. 
In this experiment, we applied our proposed \textit{C-DaSh} method and compare it against G-Shapley and TMC-Shapley approximations. For each method, we removed the lowest-quality data either chunks (for \textit{C-DaSh}) or individual tuples (for the baselines) at removal rates of $\lambda$ equal with $0.1$, $0.2$, and $0.3$. A lower absolute LOF score indicates a more effective removal of outlier data tuples. In Table \ref{tab:table-outlier}, bold values represent the best-performing method for each dataset and $\lambda$ value. Our proposed method consistently outperformed both of G-Shapley and TMC-Shapley across all datasets. This demonstrates that chunk-based subset selection in \textit{C-DaSh} provides a more robust mechanism for identifying and removing low-quality or outlier data. Also, is notably at $\lambda$ equal with $0.1$, \textit{C-DaSh} achieved the lowest absolute LOF score for all the datasets, suggesting that most outliers are captured early in the removal process. As $\lambda$ increases, the absolute LOF score increases, indicating that some high-quality tuples are removed. This suggests that the most uninformative or noisy data are identified and eliminated very early.
Additionally, compared to G-Shapley, which also leverages information from SGD for the Shapley value estimation, \textit{C-DaSh} consistently demonstrates superior with chunk-based evaluation strategy.

\subsubsection{Speedup Evaluation}

\begin{table}[!htpb]
    \centering
    \setlength\doublerulesep{0.5pt}
    \caption{C-DaSh Speedup over the G-Shapley and TMC-Shapley computational efficiency}
    \label{tab:table-speedup}
    \begin{tabular}{||c|c|c|c||}
        \hline
         \makecell{Dataset\\Name}& \makecell{Chunk Size} & \makecell{Speedup\\over\\G-Shapley ($\times$)} & \makecell{Speedup\\over\\TMC-Shapley ($\times$)}\\ \hline\
          \multirow{6}{*}{Adult} & $50$ & $23.6$ & $413.11$ \\ \cline{2-4}
           & $100$ & $31.76$ & $555.88$ \\ \cline{2-4}
            & $150$ & $41.14$ & $720$ \\ \cline{2-4}
            & $250$ & $\mathbf{108}$ & $\mathbf{1890}$ \\ \cline{2-4}
            & $500$ & $25.71$ & $450$ \\ \cline{2-4}
            & $1000$ & $24.68$ & $432$ \\ \cline{1-4}
            \multirow{6}{*}{\makecell{Bank\\Marketing}} & $50$ & $36.56$ & $618.75$ \\ \cline{2-4}
           & $100$ & $75.36$ & $1277.41$ \\ \cline{2-4}
            & $150$ & $77.6$ & $1315.37$ \\ \cline{2-4}
            & $250$ & $\mathbf{80.68}$ & $\mathbf{1365.51}$ \\ \cline{2-4}
            & $500$ & $49.26$ & $833.68$ \\ \cline{2-4}
            & $1000$ & $48.24$ & $816.49$ \\ \cline{1-4}
            \multirow{6}{*}{MIMIC-III} & $50$ & $49.31$ & $467.12$ \\ \cline{2-4}
           & $100$ & $74.22$ & $853.61$ \\ \cline{2-4}
            & $150$ & $112.5$ & $1293.75$ \\ \cline{2-4}
            & $250$ & $\mathbf{200}$ & $\mathbf{2300}$ \\ \cline{2-4}
            & $500$ & $53.73$ & $617.912$ \\ \cline{2-4}
            & $1000$ & $36.54$ & $420.30$ \\ \cline{1-4}
        \end{tabular}
\end{table}

Table \ref{tab:table-speedup} depicts the speedup achieved between \textit{C-DaSh} and baseline methods \cite{ghorbani2019data} (G-Shapley and TMC-Shapley). For each dataset, the chunk size that achieved the highest speed over the two baseline methods is highlighted in bold. Our \textit{C-DaSh} method, which uses data chunks rather than individual tuples, achieves significant speedups in computation. Finding the contribution of chunks, rather of individual data tuple, highly improves computational efficiency among all datasets. Increasing the chunk size up to $250$ data tuples leads to improved computational efficiency. This improvement occurs because fewer chunks exist and the subset selection requires less time, allowing more effective identification of high-quality data chunks that contribute positively to the learning algorithm. However, when the chunk size exceeds 250 tuples, the computational efficiency of our \textit{C-DaSh} drops, This is because chunks are more likely to contain both high-quality and bigger proportion of low-quality data tuples, making it harder to identify the highest-quality chunks during subset selection. 

\subsection{Chunk Size Evaluation}
\begin{figure}[!ht]
    \subfloat[Adult Dataset]{
        \includegraphics[width=0.35\textwidth]{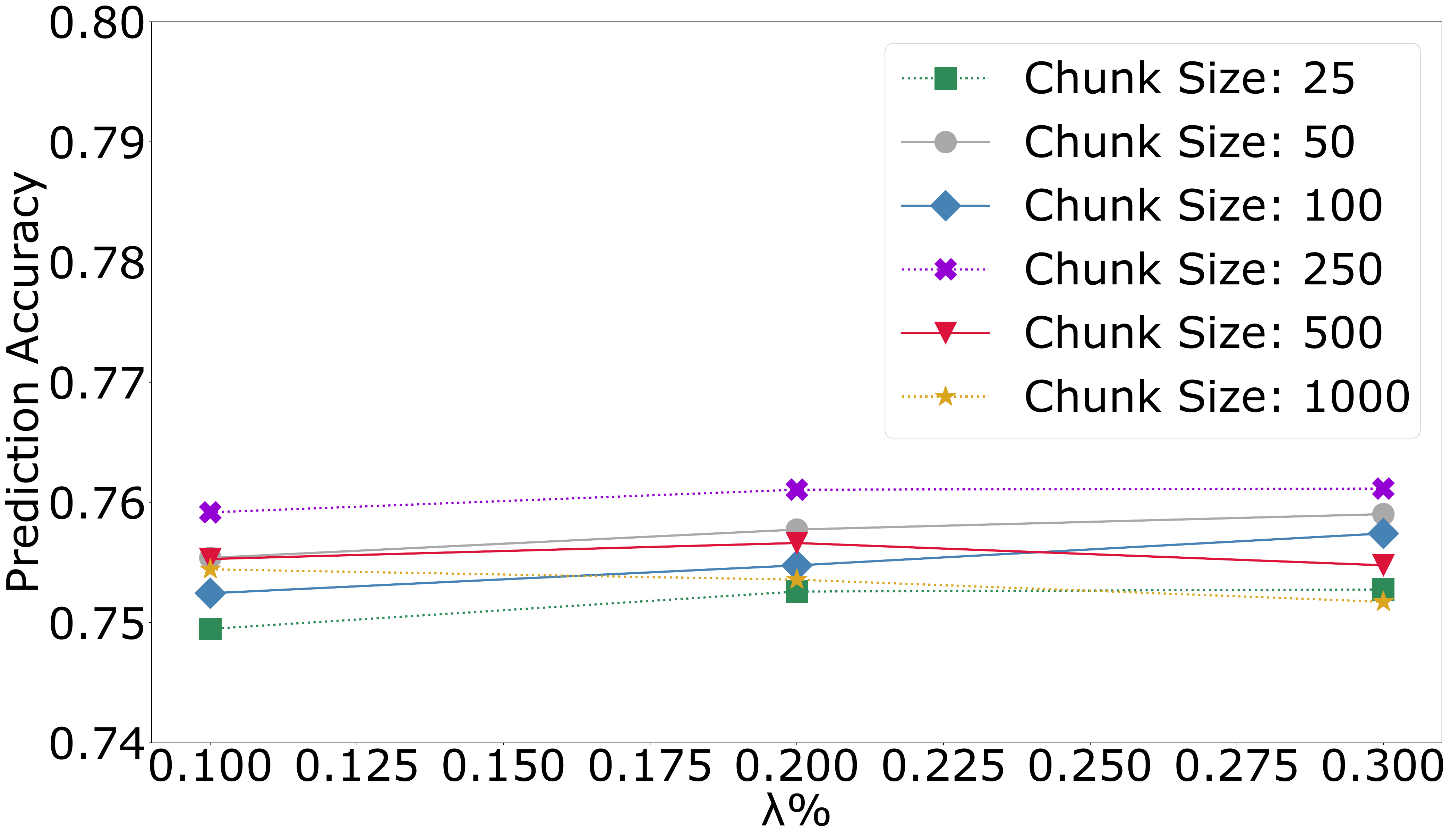}
        \label{fig:eval4-Adult}
    }
    
    \subfloat[Bank Marketing]{
        \includegraphics[width=0.35\textwidth]{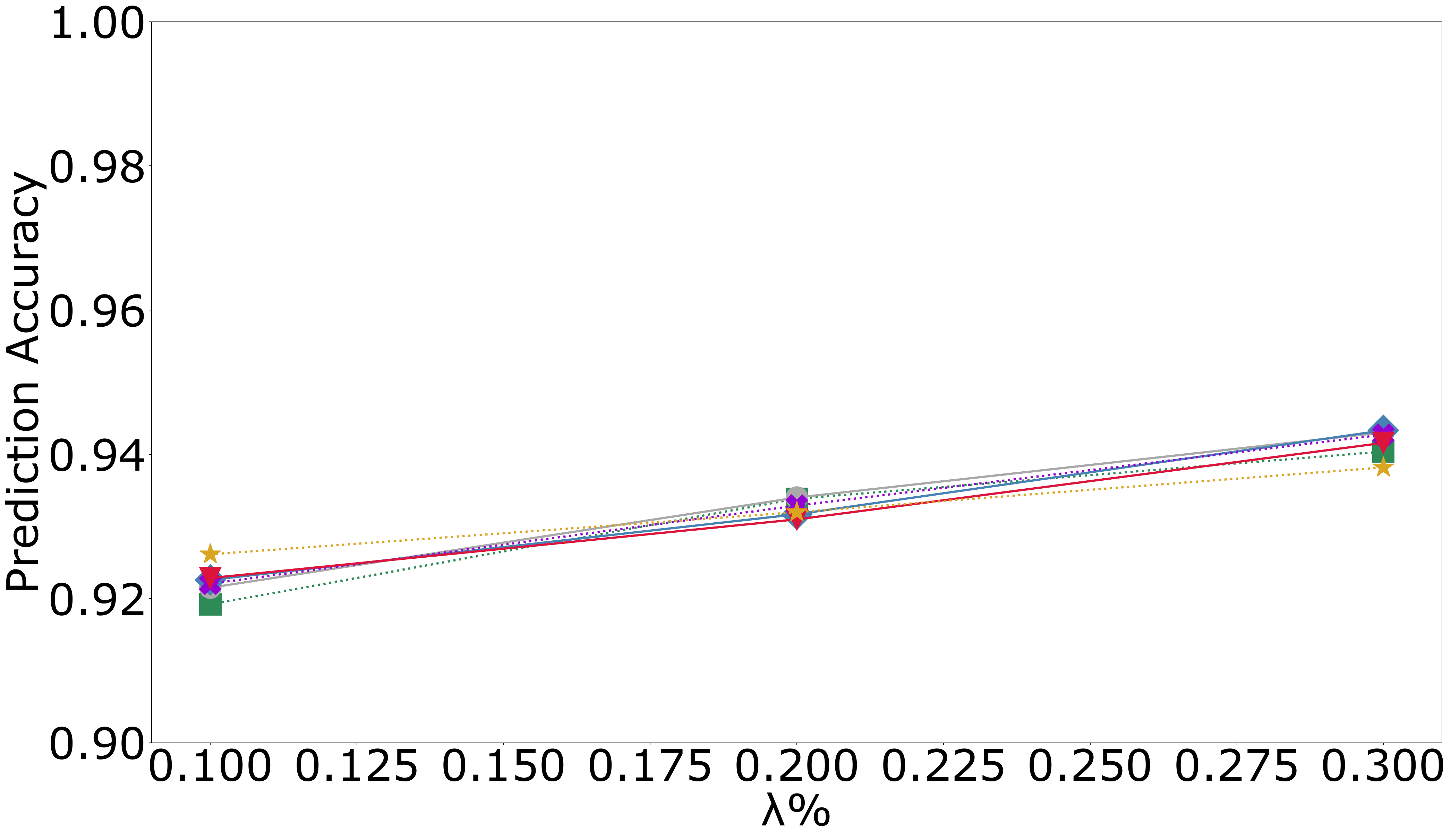}
         \label{fig:eval4-Bank}
    }
    \caption{Prediction Accuracy comparison over different chunk size by removing the $\lambda\%$ lowest data quality chunks.}

    \label{fig:eval4}
\end{figure}

In this experiment, we evaluated the impact of varying chunk sizes on the performance of our proposed approximation method, \textit{C-DaSh}, using the Adult and Bank Marketing datasets. We varied the chunk size from $50$ to $1000$ data tuples to assess its effect on the accuracy of the machine learning algorithm $\mathcal{A}$. For all the experiments, we set the number of subsets $k$ to $50$. We removed the lowest-quality data chunks, with $\lambda$ ranging from $0.1$ to $0.3$. 
Figure \ref{fig:eval4} presents the results of this experiment. Figure \ref{fig:eval4-Adult} corresponds to the Adult dataset, while Figure \ref{fig:eval4-Bank} shows results for the Bank Marketing dataset. In both sub-figures, the y-axis represents the prediction accuracy, while the x-axis shows the $\lambda$ percentage of removed low-quality data tuples. For the adult dataset (Figure \ref{fig:eval4-Adult}), we observe that for chunk sizes between $50$ and $256$ data tuples, accuracy did not significantly change as more low-quality chunks were removed. However, for chunk size equal to $256$ our method achieved its peak accuracy with a chunk size around $0.5\%$ of the total dataset size. As the chunk size increased to $500$ and $1000$ tuples, the performance of the algorithm $\mathcal{A}$ slightly drops. This declines is due to valuable data tuples being included in the removed chunks, which negatively impacts the algorithm’s learning process. For the Bank Marketing dataset (Figure \ref{fig:eval4-Bank}), at $\lambda$ equal to $0.1$, a chunk size of $1000$ tuples achieves the highest accuracy, with a small margin over the next best size. This suggests that for larger chunk sizes can effectively identifies low-quality data tuples, although the accuracy gains may be marginal compared to smaller chunks. However, the differences between the prediction accuracy and chunk sizes remains relatively small. This depicts \textit{C-DaSh}, using the subset selection, effectively identifies low-quality chunks without affecting the prediction accuracy. Based on Figure \ref{fig:eval4} and Table \ref{tab:table-speedup}, a chunk size of approximately $250$ data tuples offers the most balanced trade-off between speedup and prediction accuracy.

\subsection{Subset Size Evaluation}

\begin{figure}[!ht]
    \subfloat[Adult Dataset]{
        \includegraphics[width=0.35\textwidth]{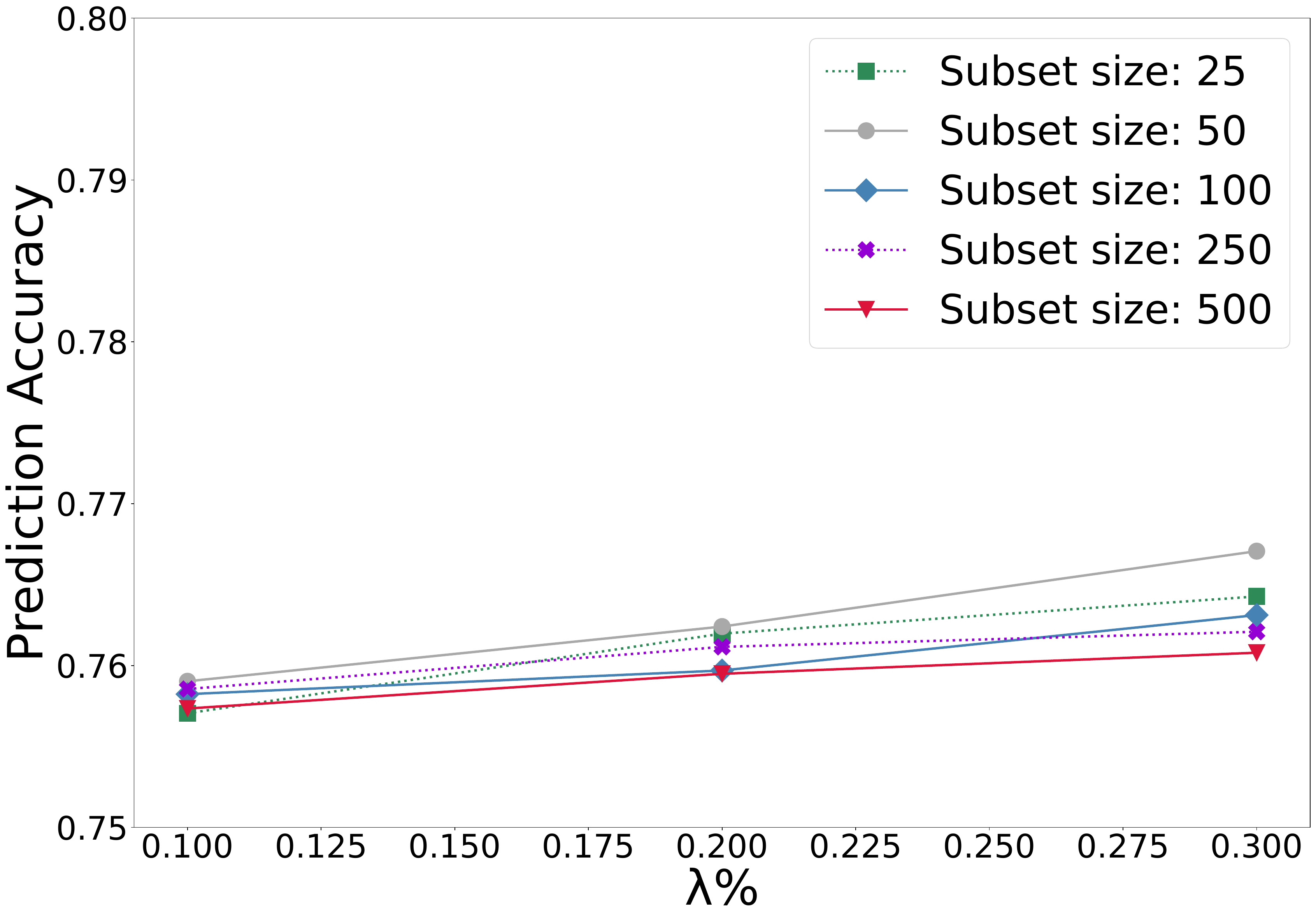}
        \label{fig:eval6-Adult}
    }
    
    \subfloat[MIMIC-III Dataset]{
        \includegraphics[width=0.35\textwidth]{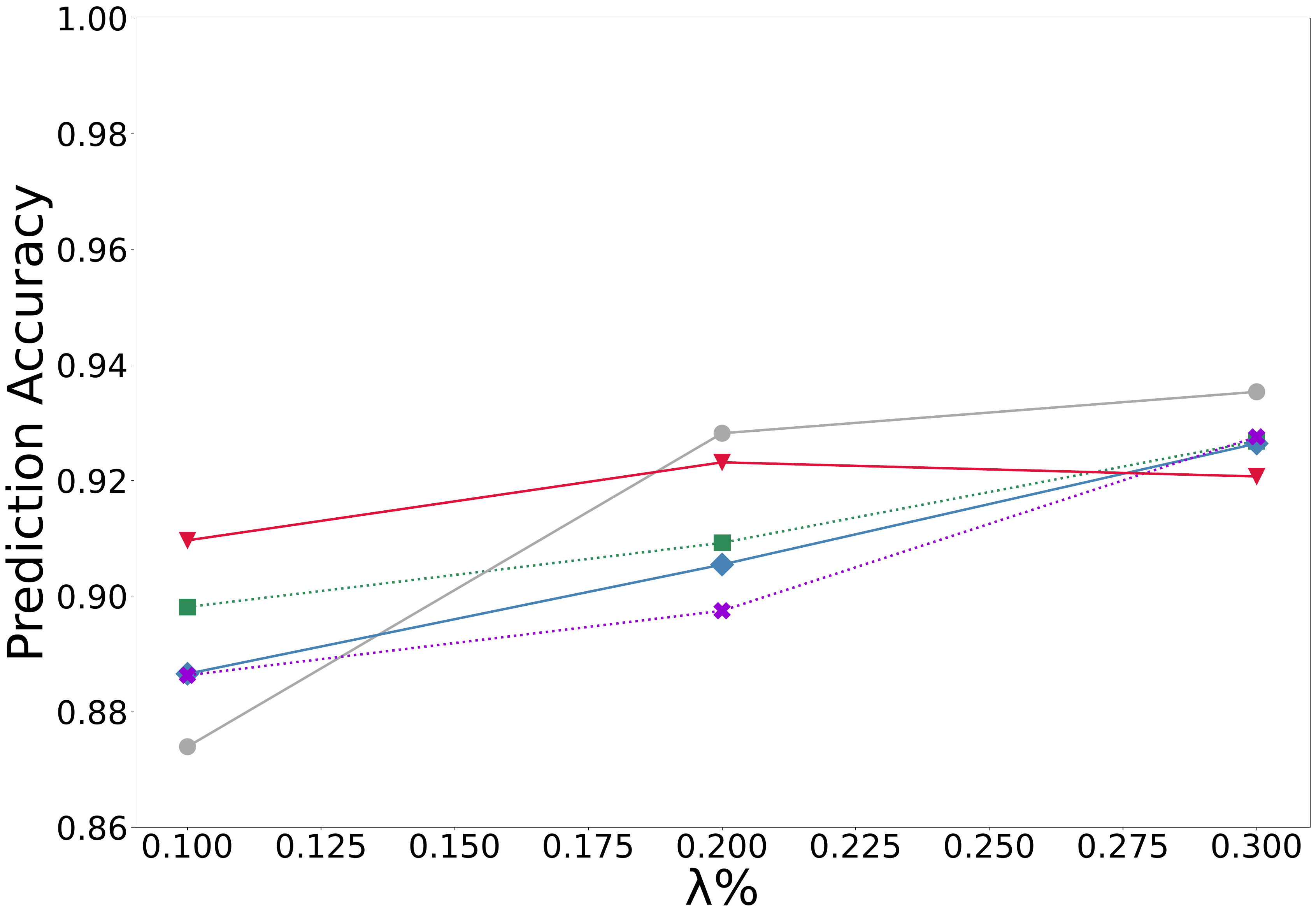}
         \label{fig:eval6-Mimic}
    }
    \caption{Prediction Accuracy comparison on \textit{C-DaSh} over different subset size by removing the $\lambda\%$ lowest data quality chunks.}
    \label{fig:eval6}
    \vspace{-0.2in}
\end{figure}

Figure \ref{fig:eval6} illustrates the experiment to evaluate \textit{C-DaSh} under varying subset sizes on the Adult (Figure \ref{fig:eval6-Adult}) and MIMIC-III (Figure \ref{fig:eval6-Mimic}). In this experiment, the subset size $k$ ranged from $25$ to $500$, while the chunk size was fixed at $256$ data tuples for both datasets, because the trade-off between the chunk size and the speedup. In both sub-figures, the x-axis represents the $\lambda\%$ of the removed low data quality chunks, and the y-axis the prediction accuracy. 
The results indicate that increasing the subset size $k$ had minimal impact to the impact on prediction accuracy for both datasets, as the accuracy remained close. This suggests that the subset selection step in \textit{C-DaSh}, effectively helps identify the most optimal subsets and improve the prediction accuracy regardless the $k$ size. Notably, a subset size $k$ of $50$ achieves the best performance, with accuracy only $0.05\%$ below the second-best result for the Adult dataset, and $1\%$ lower for MIMIC-III. Overall, these results highlight that \textit{C-DaSh} is robust across varying subset sizes and can deliver high-accuracy performance even with relatively few subsets.

\subsection{Experimental evaluation on regression datasets}
\begin{figure}[!ht]
    \subfloat[HPC Dataset]{
        \includegraphics[width=0.35\textwidth]{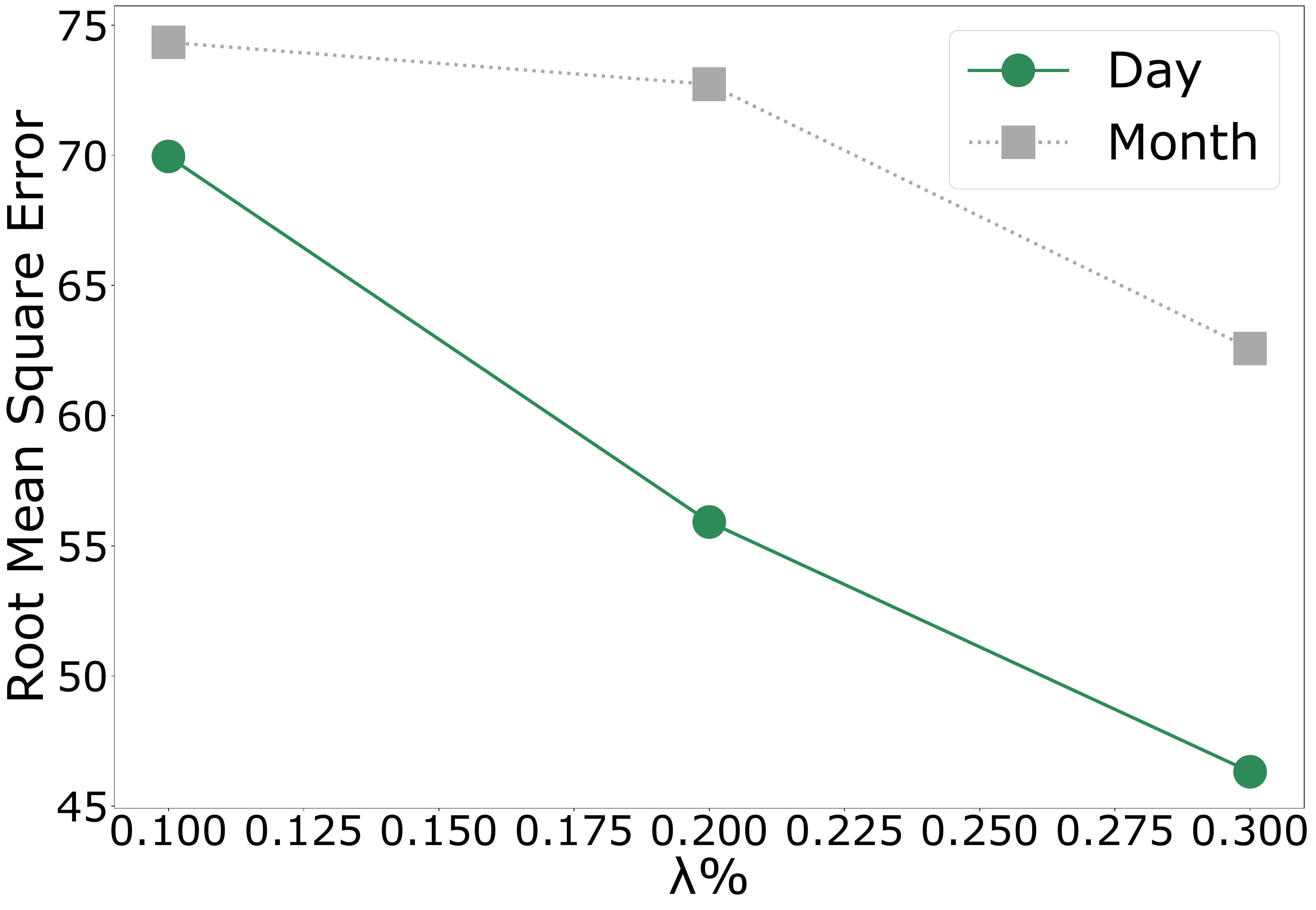}
        \label{fig:eval5-HPC}
    }
    
    \subfloat[Air Quality]{
        \includegraphics[width=0.35\textwidth]{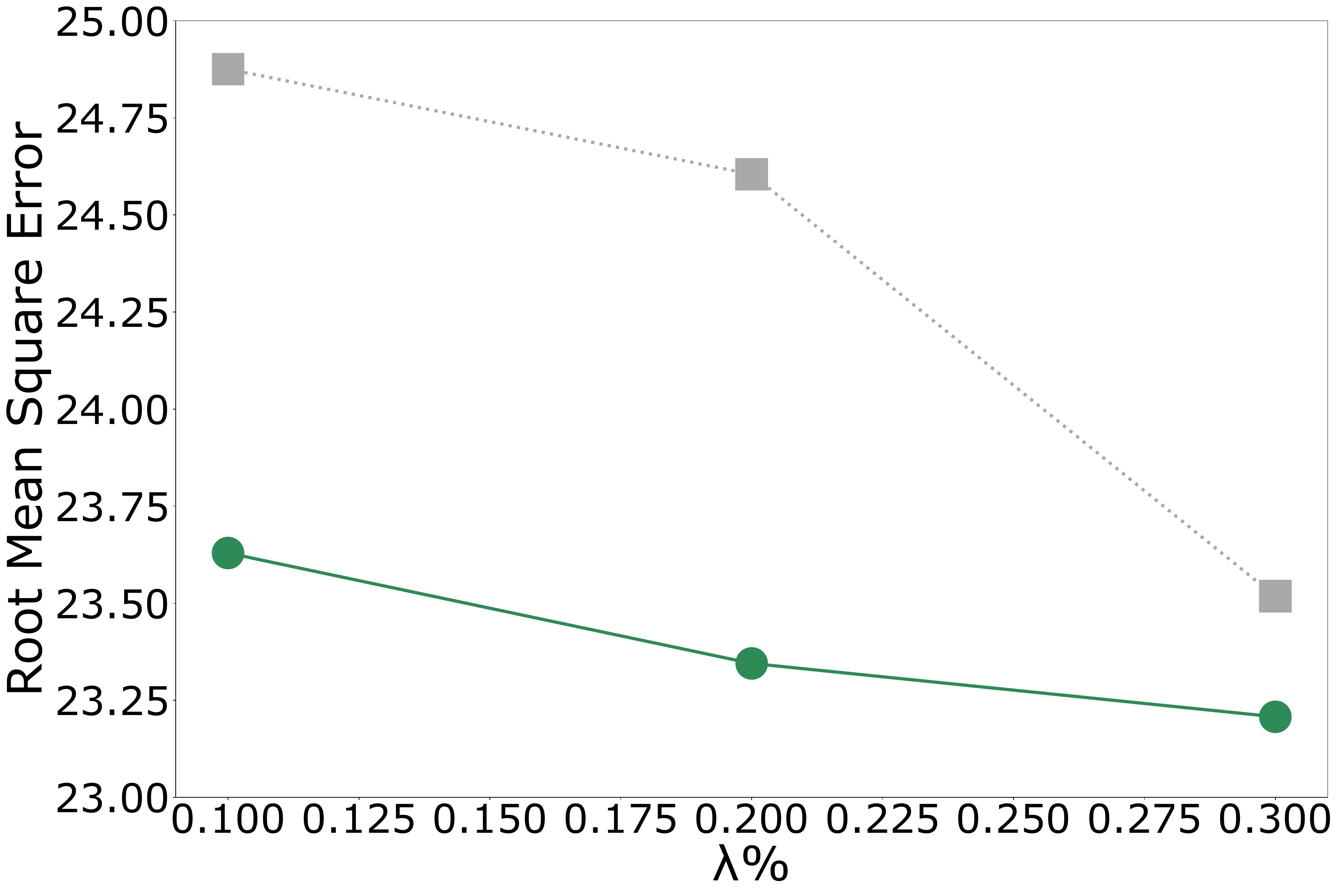}
         \label{fig:eval5-AirQuality}
    }
    \caption{\textit{C-DaSh} prediction accuracy on Regression Dataset for daily and monthly chunks by removing the $\lambda\%$ lowest data quality chunks.}
    \label{fig:eval5}
    \vspace{-0.15in}

\end{figure}

Figure \ref{fig:eval5} illustrates the performance of \textit{C-DaSh} on two regression datasets, demonstrating its effectiveness in regression tasks. For this experiment, we used Root Mean Square Error (RMSE) as the performance metric, as it effectively captures both the magnitude and direction of prediction errors. Figure \ref{fig:eval5-HPC} shows the prediction accuracy of the HPC dataset, and Figure \ref{fig:eval5-AirQuality} for the Air Quality dataset. For each dataset from the $n$ available datasets from both Air quality and HPC, we removed $\lambda\%$ of the chunks with the lowest quality value $DS_j$. In both sub-figures, the x-axis shows the percentage of removed low-quality data chunks, and the y-axis shows the RMSE (lower is better) prediction error loss function. Since regression datasets are sensitive to temporal order, we preserved this order by constructing chunks based on daily or monthly intervals, ensuring the natural structure of the data remained intact. For both experiments, the number of subsets $k$ was fixed at $50$. The results show that our method for both daily and monthly effectively identifies and removes low-quality chunks, leading to reduced RMSE as $\lambda\%$ increases. Furthermore, we observe that for a daily chunk size, our algorithm performs better and can identify the days that include missing or corrupted data. In contrast, for monthly chunks, the error is higher because removing a chunk may eliminate both low-quality and high-quality data tuples. On this experiment, we evaluate only \textit{C-DaSh}, as the baseline methods are not applicable to regression algorithms $\mathcal{A}$, with the exception of logistic regression, which is primarily used for classification tasks.

%% file: Sections/related_work.tex
The goal of data quality methods is to examine dataset content and select the ``right'' data for training a Machine Learning algorithm instead of boosting the algorithm itself. Shapley value \cite{shapley1953value} has been used to select the identical or valuable data tuples for the training of Machine Learning algorithm. However, because of the exponential computational complexity time of Shapley value calculation, approximations studies are suggested to reduce complexity time without affecting the accuracy of detecting the low-quality data tuples with traditional data errors (such as outliers, corrupted data, missing values, etc). Study from \cite{ghorbani2019data}, proposed two approximations (Truncated Monte Carlo Shapley and Gradient Shapley) to improve the identification of low-quality data tuples. Identifying the most optimal subsets to improve the quality of Shapley values has been suggested and implemented from \cite{mitchell2022sampling}. Their method uses a kernel Hilbert space in the permutation functions and checks the connections between permutations and the hypersphere, tries to identify the highest-quality subsets. On lower-dimensional problems, their method has been shown to be more effective than the higher-dimensional problems. 2-D Shapley \cite{liu20232d} is a Data Shapley approximation where, based on the two-dimensional counterfactual calculation, they abstract the block valuation problem into a two-dimensional cooperative game. Their proposed axioms for the block valuation results to a unique representation of the Shapley value for each block. In \cite{wang2024data}, noted the high complexity time issue on the Data Shapley and its approximation methods when applied to different subsets. To address this, they proposed an approximation that uses first or second-order Taylor expansions and gradient dot-products or gradient-Hessian products between training and validation data, to compute the Shapley value for each data tuple. A distributional approach to Data Shapley computation was proposed in the study by \cite{ghorbani2020distributionalframeworkdatavaluation}, where the Shapley value of a data tuple is defined with respect to an underlying data distribution. A Multi-Modal Data Shapley approach was proposed by \cite{luo2024shapley}, introducing the \textit{Shapley Value-based Contrastive Alignment} (Shap-CA) method to align image and text modalities using a Data Shapley approximation. In their framework, each image-text pair is considered as a player, and its vector embedding is extracted using a Large Language Model (LLM). Contrastive alignment is then applied to align the embeddings of the paired objects. The Shapley value measures the marginal contribution of each image-text pair to the overall vector embedding representation alignment. A higher Shapley value indicates a stronger semantic alignment and greater mutual contribution between the image and text components.

Other previous studies tried to identify low-quality data tuples without using the Data Shapley value. Commet \cite{mohammed2025step} optimises ML algorithms through a step-by-step recommendation that identify and clean low-quality data features with data errors (e.g., missing values, Gaussian noise, categorical shift, and scaling), while maximising data cleaning efficiency under resource constraints. The study by \cite{mohammed2025effects} demonstrates six data quality dimensions relevant to ML and AI algorithms and proposes various data quality measurement metrics to asses the quality of each data tuple on each of these issues. The study aims to demonstrate the relationship between the data quality dimensions and the main impact on the algorithm performance. Our work builds on the concept of Data Shapley by partitioning the dataset into equally sized chunks and computing a quality score for each chunk, rather than for each individual data tuple as done in previous Data Shapley approximation methods. Additionally, we significantly reduce computational complexity by approximating the Data Shapley values using only $k$ selected subsets of data chunks, where each chunk contains $l$ data tuples ($l \ll n$).

%% file: Sections/conclusion_and_fw.tex
In this work, we addressed the challenge of assessing the ``quality'' of a dataset's data tuple. We introduced a novel data Shapley approximation method, termed \textit{C-DaSh}, which estimates the quality of individual data chunks. By incorporating a subset selection step before the data Shapley iterations, our approach effectively identifies the most suitable quality subsets. Leveraging information retrieved from stochastic gradient descent (SGD) enhances our approximation to identify the low-quality data tuples more accurately.
Our method can be applied to both classification and regression tasks in machine learning models that use SGD as the optimisation algorithm. Empirically, we demonstrated that our approach outperforms two recent approximations, G-Shapley and TMC-Shapley \cite{ghorbani2019data}, in identifying low-quality data tuples across various data quality issues, while also achieving significant speedups from 80$\times$ to 2300$\times$. Additionally, we explored the trade-offs between execution time and accuracy over different chunk sizes, and analysed how subset size impacts the precision of detecting low-quality data.
For future work, we aim to improve our subset selection strategy by introducing a data structure that groups low-quality data tuples in the same chunks, thereby enhancing the precision of quality identification. Furthermore, we plan to extend our method to support \textit{multi-modal datasets} that combine textual descriptions (e.g., captions or subtitles) with images or videos, enabling the evaluation of data quality across different modalities.